\documentclass{article}

 \usepackage[dblblindworkshop, final]{neurips_2025}
 \usepackage{graphicx}
 \usepackage{subcaption}
 \usepackage{amsmath}
\usepackage{booktabs}   
\usepackage{multirow}   
\usepackage{makecell}

\workshoptitle{Foundation Models for the Brain and Body}

\usepackage[utf8]{inputenc} 
\usepackage[T1]{fontenc}    
\usepackage{hyperref}       
\usepackage{url}            
\usepackage{booktabs}       
\usepackage{amsfonts}       
\usepackage{nicefrac}       
\usepackage{microtype}      
\usepackage{xcolor}         

\title{CPEP: Contrastive Pose-EMG Pre-training Enhances Gesture Generalization on EMG Signals}

\author{
  Wenhui Cui$^{1,2}$\thanks{Work completed during internship at Apple.}, 
  Christopher Sandino$^{1}$, 
  Hadi Pouransar$^{1}$, 
  Ran Liu$^{1}$, 
  Juri Minxha$^{1}$, \And
  Ellen L. Zippi$^{1}$, 
  Aman Verma$^{1}$, 
  Anna Sedlackova$^{1}$, 
  Erdrin Azemi$^{1}$, 
  Behrooz Mahasseni$^{1}$ \\
  \\
  $^{1}$Apple \\
\texttt{\{csandino, mpouransari, ran\_liu3, j\_minxha, ezippi,}\\
\texttt{amanverma, asedlackova, erdrin, bmahasseni\}@apple.com} \\
  $^{2}$Ming Hsieh Department of Electrical and Computer Engineering, University of Southern California \\
  \texttt{wenhuicu@usc.edu}
}

\begin{document}

\maketitle

\begin{abstract}

Hand gesture classification using high-quality structured data such as videos, images, and hand skeletons is a well-explored problem in computer vision. Leveraging low-power, cost-effective biosignals, e.g. surface electromyography (sEMG), allows for continuous gesture prediction on wearables. In this paper, we demonstrate that learning representations from weak-modality data that are aligned with those from structured, high-quality data can improve representation quality and enables zero-shot classification.
Specifically, we propose a \textbf{C}ontrastive \textbf{P}ose-\textbf{E}MG \textbf{P}re-training (\textbf{CPEP}) framework to align EMG and pose representations, where we learn an EMG encoder that produces high-quality and pose-informative representations.
We assess the gesture classification performance of our model through linear probing and zero-shot setups. Our model outperforms \textit{emg2pose} benchmark models by up to \textbf{21\%} on in-distribution gesture classification and \textbf{72\%} on unseen (out-of-distribution) gesture classification.



\end{abstract}

\section{Introduction}
Biosignals, such as inertial measurement unit (IMU) and surface electromyography (sEMG), have been widely applied in rehabilitation, health monitoring, human activity recognition, and robotic control due to their low power requirements and ease of integration into wearable devices \cite{khedr2017smartphone, jarque2021systematic, gu2023imu, haresamudram2025past}. Human gesture recognition on wearables has recently attracted significant interest and demonstrated potential across various scenarios \cite{pyun2024machine, wang2023mems, moin2021wearable}. However, predicting hand gestures from biosignals like sEMG, especially generalizing to unseen gestures, remains challenging \cite{10.1145/3290605.3300568}, primarily because of the high variability and fine dexterity of human hand movements, sensor noise, and/or the limited scale of publicly available data.
Thanks to recent advances in deep learning and the availability of large-scale visual data, such as videos and motion capture data \cite{casile2023quantitative}, vision-based models have achieved remarkable success \cite{pavlakos2023reconstructinghands3dtransformers}. Despite these advances, cameras’ high power demands and privacy concerns hinder deployment on wearables, and pose occlusions can destabilize vision-based gesture classification. This has driven a recent rise in the use of low-power sensors, such as IMUs and EMGs, for real-time gesture recognition \cite{dahiya2024efficient, tchantchane2023review}. However, collecting large-scale datasets using biosignals in unconstrained environments remains challenging and expensive \cite{sivakumar2025emg2qwertylargedatasetbaselines}. To overcome this, we introduce a representation alignment approach that allows the model to generalize to unseen gestures and new users at test time without the need for additional training or extensive data collection.



Through our experiments, we found that due to the inherently noisy nature of EMG signals, conventional self-supervised pre-training methods are suboptimal and fail to achieve high-quality gesture classification. We also found that supervised training approaches, which directly predict poses from EMG signals \cite{salter2024emg2poselargediversebenchmark}, have limited generalization capacity to unseen gestures or users. Therefore, we explore a contrastive pre-training strategy in this work.
Aligning image–text representations through contrastive pre-training has been shown to enhance the quality of both text and image embeddings \cite{radford2021learningtransferablevisualmodels, li2023blip2}. Unlike CLIP \cite{radford2021learningtransferablevisualmodels} model, which is trained on billions of image–text pairs, we align pre-trained unimodal encoders to reduce the amount of paired data required. We introduce \textbf{C}ontrastive \textbf{P}ose-\textbf{E}MG \textbf{P}re-training (\textbf{CPEP}), a method that enables zero-shot classification of new gestures without the need for task- or data-specific fine-tuning. Specifically, we use two encoders: (a) an EMG encoder, pre-trained on unlabeled EMG signals via self-supervised learning, and (b) a pose encoder, pre-trained on unlabeled pose data to produce high-quality pose representations. A contrastive learning objective then aligns the EMG representation with its corresponding pose representation in the embedding space, while separating mismatched pairs. CPEP encourages the EMG encoder to capture pose-relevant information and develop a rich understanding of diverse hand kinematics, enabling generalization to unseen gestures by extrapolating within the structured latent space. In CPEP, 
pose encoder provides richer supervisory signals by capturing structural and semantic relationships that are difficult to learn from the weak modality (EMG) alone, thus guiding the weak-modality encoder toward more discriminative and generalizable representations.

We utilize the large-scale public EMG dataset \textit{emg2pose} \cite{salter2024emg2poselargediversebenchmark}, which provides simultaneous paired EMG and pose recordings 
to pre-train our CPEP model and perform downstream evaluations. Leveraging the metadata available in the emg2pose dataset, we design a gesture classification task to evaluate EMG representation quality. Following CLIP evaluation protocol \cite{radford2021learningtransferablevisualmodels}, we validate the learned EMG representations through zero-shot classification and linear probing, demonstrating superior performance on both in-distribution and unseen gestures compared to \textit{emg2pose} benchmark models.
In summary, our contributions are two-fold:
(i) We propose a contrastive pre-training strategy to enhance EMG representation learning from noisy EMG signals by aligning them with high-quality pose representations.
(ii) To the best of our knowledge, CPEP is the first framework enabling zero-shot classification of unseen gestures from EMG signals.

\section{Methods}
\begin{figure}
  \centering
  \includegraphics[width=0.95\linewidth]{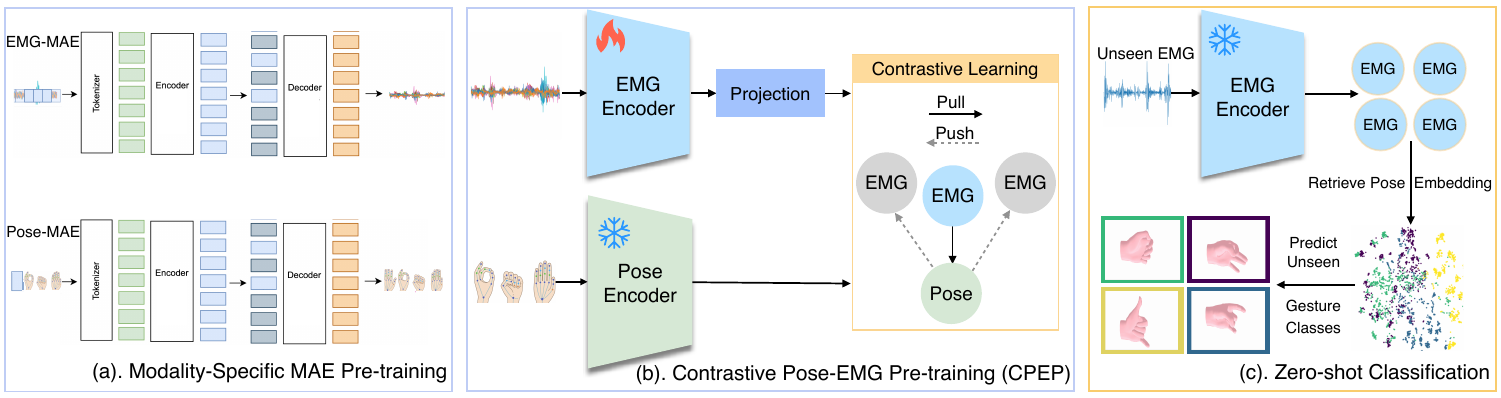}
 \caption{Overview of Contrastive Pose–EMG Pre-training (CPEP) framework. (a). MAE pre-trains EMG and pose encoders; (b). CPEP freezes the pose encoder and aligns EMG to pose via contrastive learning, enabling (c). zero-shot gesture classification by retrieval over frozen pose embeddings.
}
  \label{fig:clip}
\end{figure}

In this section, we formalize the problem, describe the pre-training of unimodal EMG and pose encoders, and present our contrastive pre-training strategy CPEP for aligning Pose-EMG representations. An overview of the full pipeline is shown in Fig.~\ref{fig:clip}.


\newtheorem{myDef}{Definition}
\subsection{Problem Definition}

\begin{myDef}\textbf{EMG, Pose data, and \textbf{Gesture classes.} } Let $\mathcal{X}=\{\mathbf{x}_i\in\mathbb{R}^{C\times T}\}_{i=1}^{N}$ denote multi-channel EMG sequences with $C$ channels and time length $T$, and let $\mathcal{P}=\{\mathbf{p}_i\in\mathbb{R}^{J\times T}\}_{i=1}^{N}$ denote the paired pose sequences of joint angles, where $J$ is the number of joints in a pre-defined hand-skeleton \cite{salter2024emg2poselargediversebenchmark}. Here, $N$ is the total number of paired samples.
 
Let $\mathcal{Y}=\{1,\dots,K\}$ denote the gesture classes, and let $\{y_i\}_{i=1}^N$ be the labels with $y_i\in\mathcal{Y}$. Each pose $p_i$ has a unique label $y_i$, although multiple samples may share the same label. The paired dataset is
\[
\mathcal{D}=\{(x_i,p_i,y_i)\}_{i=1}^N,
\]
where $x_i$ and $p_i$ are a paired EMG–pose sample and $y_i$ is its gesture label. We split $\mathcal{D}$ into $\mathcal{D}_{\mathrm{tr}}$, $\mathcal{D}_{\mathrm{val}}$, and $\mathcal{D}_{\mathrm{test}}$. Let $\mathcal{Y}_{\mathrm{in}}\subset\mathcal{Y}$ be the in-distribution gesture classes and $\mathcal{Y}_{\mathrm{unseen}} \subset\mathcal{Y}$ be the unseen gesture classes, where $\mathcal{Y}_\mathrm{in} \cap \mathcal{Y}_{\mathrm{unseen}} = \emptyset$. For $S\in\{\mathrm{tr},\mathrm{val},\mathrm{test}\}$,
\[
\mathcal{D}_S^{\mathrm{in}}=\{(x,p,y)\in\mathcal{D}_S:\,y\in\mathcal{Y}_{\mathrm{in}}\},\qquad
\mathcal{D}_S^{\mathrm{unseen}}=\{(x,p,y)\in\mathcal{D}_S:\,y\in\mathcal{Y}_{\mathrm{unseen}}\}.
\]

\end{myDef}

\begin{myDef}\textbf{Encoders.} Let $\mathcal{E}_x$ and $\mathcal{E}_p$ denote encoder functions mapping EMG and pose data respectively into latent embeddings in $\mathbb{R}^{d}$.\end{myDef}


\subsection{Unimodal Encoder Pre‐training}

We adopt the standard Transformer encoder \cite{vaswani2023attentionneed} backbone with a linear tokenizer to map raw signals into $d$‐dimensional token embeddings. A patch length of $S$ along the temporal dimension results in $K = \left\lfloor \frac{T}{S} \right\rfloor$ non‐overlapping tokens. We flatten the channels at each time-point within the patch and use linear tokenizer to project each patch into an embedding $\mathbf{z_i} \in \mathbb{R}^d$. 
Following Masked autoencoder (MAE) \cite{he2021maskedautoencodersscalablevision}, given a mask ratio $r \in (0, 1)$, we sample a random mask set 
$\mathcal{M}\subset\{1,\dots,K\}, |\mathcal{M}| = K \times r$. The encoder $\phi$ processes only the unmasked tokens
$\{\mathbf{z}_i : i\notin\mathcal{M}\}$,
and the decoder $\psi$ reconstructs the full sequence. We optimize the masked‐token reconstruction via mean squared error:
$
\mathcal{L}_{\mathrm{MAE}}
=\;\frac{1}{|\mathcal{M}|}
\sum_{i\in\mathcal{M}}
\bigl\|
\psi\bigl(\phi(\{\mathbf{z}_j\}_{j\notin\mathcal{M}})\bigr)_i 
\;-\;\mathbf{z}_i
\bigr\|_2^2.
$
We apply this pre-training independently to the EMG and pose data, yielding two pre-trained encoders
$\mathcal{E}_x=\phi_{\text{EMG}}$ and $\mathcal{E}_p=\phi_{\text{pose}}$,
which produce meaningful representations for contrastive alignment.



\subsection{Contrastive Pose-EMG Pre-training (CPEP)}

In our approach, the hand pose data,  offering high-quality, low-noise gesture information, serves as the anchor modality. In contrast, EMG signals exhibit lower signal-to-noise ratio (SNR). To align EMG and pose representations, we append a lightweight projection head $h_\phi$ to the pre-trained EMG encoder $\mathcal{E}_x$, mapping its [CLS] token into the same $d$-dimensional space as the pose encoder’s [CLS] token. The [CLS] token is chosen for its ability to capture global context and provide a compact summary of the entire sequence \cite{he2021maskedautoencodersscalablevision}. During contrastive training, we keep the pose encoder $\mathcal{E}_p$ frozen and only update the EMG encoder $\mathcal{E}_x$ (together with the projection head $h$) to avoid the pose representations being altered when enforcing alignment with EMG representations.
For each paired sample $(x_i,p_i)$ we compute $u_i = h\!\left(\mathcal{E}_x(x_i)\right)_{\mathrm{[CLS]}}$,
$v_i = \left(\mathcal{E}_p(p_i)\right)_{\mathrm{[CLS]}}$,
where the projection head $h$ is randomly initialized and trained, and the pose encoder $\mathcal{E}_p$ is frozen. We $\ell_2$-normalize embeddings and use cosine similarity. Let
$
\tilde u_i=\frac{u_i}{\|u_i\|},\quad \tilde v_j=\frac{v_j}{\|v_j\|},\quad
s_{ij}=\frac{\tilde u_i^\top \tilde v_j}{\tau},
$
with temperature $\tau$. Following CLIP \cite{radford2021learningtransferablevisualmodels}, the symmetric Information Noise-Contrastive Estimation (InfoNCE) objective \cite{oord2019representationlearningcontrastivepredictive} is:
\[
\mathcal{L}_{\mathrm{CPEP}}
=\frac{1}{2N}\sum_{i=1}^N\Bigg[
-\log\frac{\exp(s_{ii})}{\sum_{j=1}^N \exp(s_{ij})}
\;-\;
\log\frac{\exp(s_{ii})}{\sum_{j=1}^N \exp(s_{ji})}
\Bigg].
\]
This loss pulls EMG embedding $\tilde u_i$ toward its matching pose $\tilde v_i$ while pushing away all non-matching poses, yielding pose-informed EMG embeddings that support generalizable gesture classification.
The embedding dimension is $256$ and CPEP was trained with a batch size of $256$.

\subsection{Evaluation Protocols}
We employ two evaluation protocols to examine the learned representation quality. \textbf{Linear probing} (LP):  with labeled EMG data $\{(x_i,y_i)\}$, we freeze $\mathcal{E}_{x}^{*}$ and train a randomly-initialized linear classifier $\mathcal{C}$ on top of its embeddings, reporting accuracy on a held-out split. 
\textbf{Zero‑shot classification} (ZS) is performed as k-nearest-neighbor voting in the embedding space, following standard practice in representation learning \citep{marks2024closerlookbenchmarkingselfsupervised, radford2021learningtransferablevisualmodels}. For each EMG sample, we retrieve its TopK nearest poses in the embedding space, then vote the corresponding gesture labels to determine the predicted gesture. Given a test EMG sample $x_j$, let $
\mathcal{R}_j = \operatorname*{TopK}_{p\in\mathcal{P}_{\mathrm{test}}}
\!\bigl(\mathcal{E}_{x}(x_j)^\top  \mathcal{E}_{p}(p))\bigr)
$
be the set of $k$ pose samples with highest cosine similarity to $x_j$. We then predict 
$\hat y_j = \mathop{\mathrm{mode}}\bigl\{\,y(p)\mid p\in\mathcal{R}_j\bigr\},
$
where $y(p)$ denotes the gesture class associated with pose $p$.

\section{Experiments and Results}
\subsection{Data}

\textbf{Dataset: }We use \textit{emg2pose} \cite{salter2024emg2poselargediversebenchmark}, a large-scale open-source EMG dataset containing 370 hours of sEMG and synchronized hand pose data across 193 consenting users, 29 different behavioral groups that include a diverse range of discrete and continuous hand motions such as making a fist or counting to five. The hand pose labels are generated using a high-resolution motion capture system. The full dataset contains over 80 million pose labels and is of similar scale to the largest computer vision equivalents.
Each user completed four recording sessions per gesture category, each with different EMG band placement. Each session lasted 45–120 s, during which users repeatedly performed a mix of 3–5 similar gestures or unconstrained freeform movements. We use non-overlapping 2-second windows as input sequences. EMG is instance-normalized, band-pass filtered (2–250\,Hz), and notch-filtered at 60\,Hz.  
For more details, please refer to \cite{salter2024emg2poselargediversebenchmark}.

\begin{table}[tb]
\centering
\footnotesize
\resizebox{0.5\textwidth}{!}{%
\begin{tabular}{l l c c c c}
\toprule
\multirow{2}{*}{Split (totals)} & \multirow{2}{*}{Subset} & \multicolumn{2}{c}{Gesture Counts} & \multicolumn{2}{c}{User Counts} \\
\cmidrule(lr){3-4} \cmidrule(lr){5-6}
 &  & In-dist. & Unseen & In-dist. & Unseen \\
\midrule
\makecell{$\mathcal{D}_{tr}$ \scriptsize(23 gestures / 158 users)} & $\mathcal{D}_{\text{probe-tr}}^{\text{in}}$  & 4 & 0 & 158 & 0 \\
\midrule
\multirow{2}{*}
{\makecell{$\mathcal{D}_{val}$ \scriptsize(29 gestures / 15 users)}} 
  & $\mathcal{D}_{\text{tune}}$                    & 4 & 4 & 0 & 3 \\
  & $\mathcal{D}_{\text{probe-tr}}^{\text{unseen}}$         &  0 & 4 & 0 & 12 \\
\midrule
\multirow{1}{*}{\makecell{$\mathcal{D}_{test}$ \scriptsize(29 gestures / 158 users)}} 
  & $\mathcal{D}_{\text{eval}}$             &  4 & 4 & 0 & 20 \\
\bottomrule
\end{tabular}%
}
\caption{Dataset splits with gesture and user counts. Four unseen gestures evaluated out of six total.}
\label{tab:dataset_splits}
\end{table}

\textbf{Experiment Design:}
We evaluate on two disjoint gesture sets drawn from the public \textit{emg2pose} corpus. First, we select four representative single‑hand motions covering various finger movements as our \textbf{in‑distribution (in-dist.) gestures}. Second, from the six held‑out classes that are not seen during training, we exclude the two‑handed gesture and the highly variable ``finger freeform'' class, yielding four \textbf{unseen gestures}. Note that we include gestures like ``\textit{CountingWiggling}'' and ``\textit{FingerPinches}'' that are known to be challenging for vision-based hand pose estimation due to visual occlusion \cite{salter2024emg2poselargediversebenchmark}. Details of gesture classes are in Appendix A. 
For data splits, we follow the public train $\mathcal{D}_{tr}$, val $\mathcal{D}_{val}$ , test $\mathcal{D}_{test}$ splits in \cite{salter2024emg2poselargediversebenchmark} and define our data splits for downstream gesture classification tasks as shown in Table \ref{tab:dataset_splits}. The model is pre-trained on the full $\mathcal{D}_{tr}$. A linear head is trained on $\mathcal{D}_{\text{probe-tr}}^{\text{in}}$, and report final accuracy on the evaluation set $\mathcal{D}_{\text{eval}}^{\text{in}}$. For unseen gestures (which appear only in the original val and test splits), we train on $\mathcal{D}_{\text{probe-tr}}^{\text{unseen}}$, and report accuracy on $\mathcal{D}_{\text{eval}}^{\text{unseen}}$. 
Zero-shot classification is evaluated only on $\mathcal{D}_{\text{eval}}$. All users in $\mathcal{D}_{\text{eval}}$ are unseen, so the results also assess user-level generalization. A held-out dataset $\mathcal{D}_{\text{tune}}$, strictly disjoint from all other sets, is reserved for hyper-parameter tuning.
\begin{figure}[hbt]
  \centering
  \includegraphics[width=0.9\linewidth]{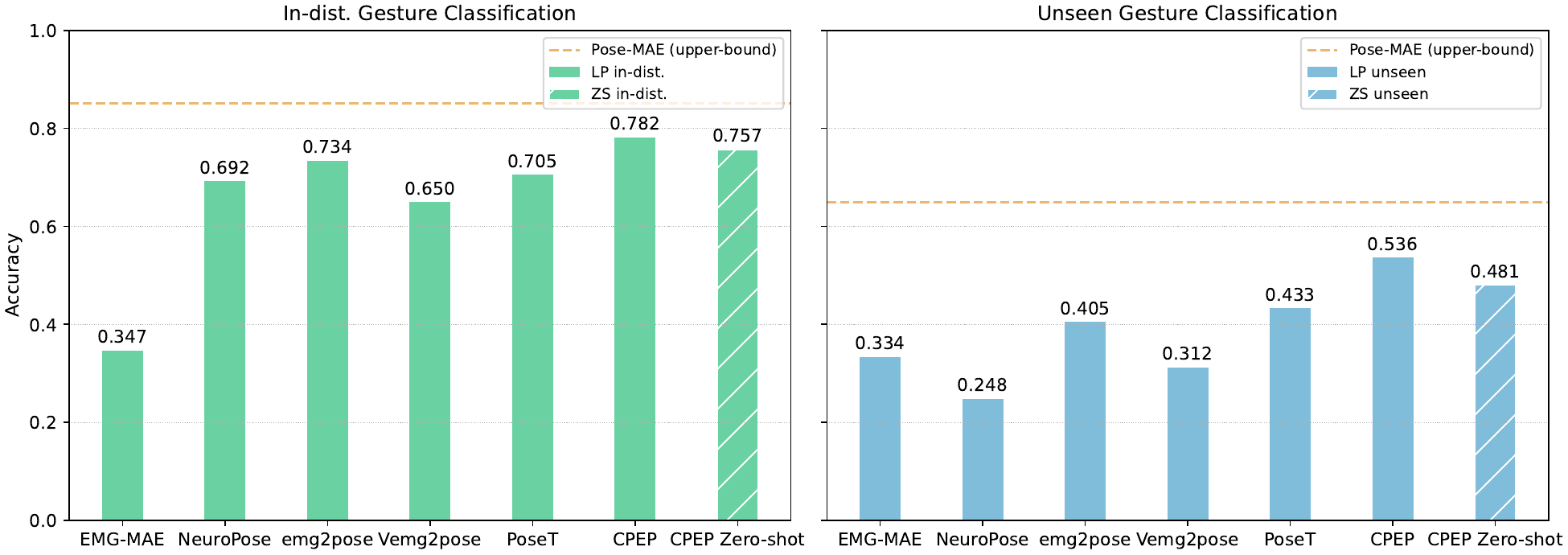}
  
\caption{Classification accuracy on in-distribution and unseen gestures. Linear probing is reported for all methods; zero-shot is reported only for CPEP (other methods do not support zero-shot).
}
 
  \label{fig: res}
\end{figure}

\subsection{Gesture Classification Results}
We assess the quality of EMG representations on both \textbf{in-dist.} and \textbf{unseen} gestures using the two evaluation protocols described earlier: zero-shot classification (ZS) and linear probing (LP). For comparative analysis, 
we compare our CPEP model against: (i) a supervised encoder–decoder Transformer called \textbf{PoseT} trained to regress poses directly from EMG; (ii) the supervised pose regression models (\textbf{emg2pose}, \textbf{Vemg2pose}, and \textbf{NeuroPose}) released with the \textit{emg2pose} benchmark \cite{salter2024emg2poselargediversebenchmark}; and (iii) a self-supervised MAE model pre-trained using EMG data only (\textbf{EMG-MAE}). For a fair comparison under linear probing, we freeze each baseline’s encoder and train a softmax linear head on top of its embeddings. For context, we include an \textbf{upper-bound} by linear probing Pose-MAE features, since the pose encoder serves as the alignment anchor. 

Figure~\ref{fig: res} presents a summary of the experimental results. Since supervised baselines do not support zero-shot (k-nearest neighbor) classification in the embedding space, we report only their LP results. We report macro accuracy (mean per-class accuracy) to account for class imbalance across gestures. We omit standard deviations over random seeds because they were negligible relative to the performance gains.  
CPEP outperforms all baselines in terms of accuracy under both linear probing and zero-shot settings. Remarkably, zero-shot classification on unseen gestures achieves performance comparable to linear probing, without requiring any additional task-specific training. Furthermore, our zero-shot CPEP approach exceeds the linear probing results of the baselines, highlighting its superior generalizability through contrastive learning. These findings support our claim that aligning EMG signals with high-quality pose embeddings enables effective generalization to unseen gestures.

\subsection{Ablation Study}

\textbf{Impact of CPEP design choices.}
Table~\ref{tab:ablation} summarizes the ablation results for different CPEP design choices. Initializing the EMG encoder from the pre-trained MAE consistently outperforms random initialization (\emph{CPEP-$\mathcal{E}_x$ RandInit}) and results in a more stable contrastive training. Random initialization slows convergence and degrades both zero-shot and linear-probing performance. One important finding is that randomly initializing both encoders ($\mathcal{E}_x$ and $\mathcal{E}_p$) fails to converge. Freezing both encoders and training only the projection head (\emph{CPEP-$\mathcal{E}_x$ Frozen}) leads to significantly lower accuracy, highlighting the necessity of fine-tuning the EMG encoder during the feature alignment phase. Making the pose encoder trainable alongside the EMG encoder (updating both) also fails to converge, suggesting that altering the pose anchor harms alignment. Furthermore, average pooling over tokens (\emph{CPEP-AvgPool}) underperforms the \texttt{[CLS]} token.

\begin{table}[htb]
    \centering
    \begin{tabular}{l|c|c|c|c}
    \toprule
      & LP in-dist. & ZS in-dist. & LP unseen & ZS unseen \\
    \midrule
    CPEP-$\mathcal{E}_{x}$ Frozen & 0.372 & 0.344 & 0.326 & 0.298 \\
    CPEP-$\mathcal{E}_{x}$ RandInit & 0.748 & 0.701 & 0.479 & 0.454 \\
    CPEP-AvgPool & 0.761 & 0.711 & 0.518 & 0.454\\
    CPEP (ours) & \textbf{0.782} & \textbf{0.757} & \textbf{0.536} & \textbf{0.481}\\
    \bottomrule
    \end{tabular}
    \caption{Ablations of CPEP training setup. }
    \label{tab:ablation}
\end{table}

\begin{figure}[htb]
    \centering
    \includegraphics[width=0.85\linewidth]{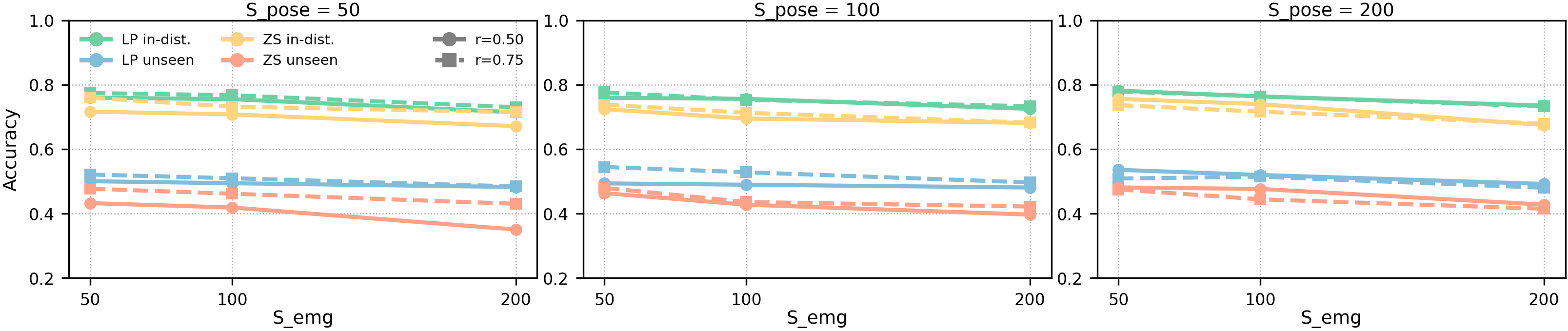}
    \caption{Ablation of MAE pre-training setup: temporal patch length $S$ and mask ratio $r$.}
    \label{fig:ablation}
\end{figure}

\textbf{Impact of $S_{\text{emg}}$, $S_{\text{pose}}$, and mask ratio $r$.}
We study the effect of the unimodal MAE pre-training choices (patch length $S$ and mask ratio $r$) on downstream classification, with CPEP training setup fixed. Fig.~\ref{fig:ablation} summarizes the results across different hyper-parameter settings. Overall, for small values of $S_{\text{pose}}$, higher $r$ values achieve better results. With $S_{\text{pose}}{=}200$, $r{=}0.5$ outperforming others. 
Using longer EMG patches consistently degrades performance.

\section{Discussion and Conclusion}
We introduced CPEP, a contrastive pre-training framework that aligns EMG with pose representations. By leveraging pose as a rich supervisory signal, CPEP learns pose-informed EMG embeddings that capture structural and semantic relationships and are linearly separable in the latent space. Across in-distribution and unseen gesture classification tasks, CPEP demonstrated strong generalization and zero-shot capabilities. 
 Future work includes extending CPEP to additional EMG datasets and other biosignals (e.g. IMU), as well as exploring tasks such as human activity recognition. 
 Overall, CPEP provides an effective approach to zero-shot gesture classification from EMG, and can serve as a foundation for broader multi-modal biosignal applications.

\bibliographystyle{plain}
\bibliography{refs}

\appendix

\section{Details of Gestures}

\paragraph{In-distribution (4 classes).}
 (1) Thumb swipes whole hand; (2) Hand claw, grasp, and flicks; (3) Thumbs rotation, thumbs up and down; (4) finger pinches, single/multiple fingers.

\paragraph{Unseen (4 classes).}
(1) Hook 'em Horns, OK, and Scissors; (2) Shaka and Vulcan peace; (3) Counting up/down; (4) Counting up/down with finger wiggling and spreading.

\begin{figure}[htb]
    \centering
    \includegraphics[width=0.95\linewidth]{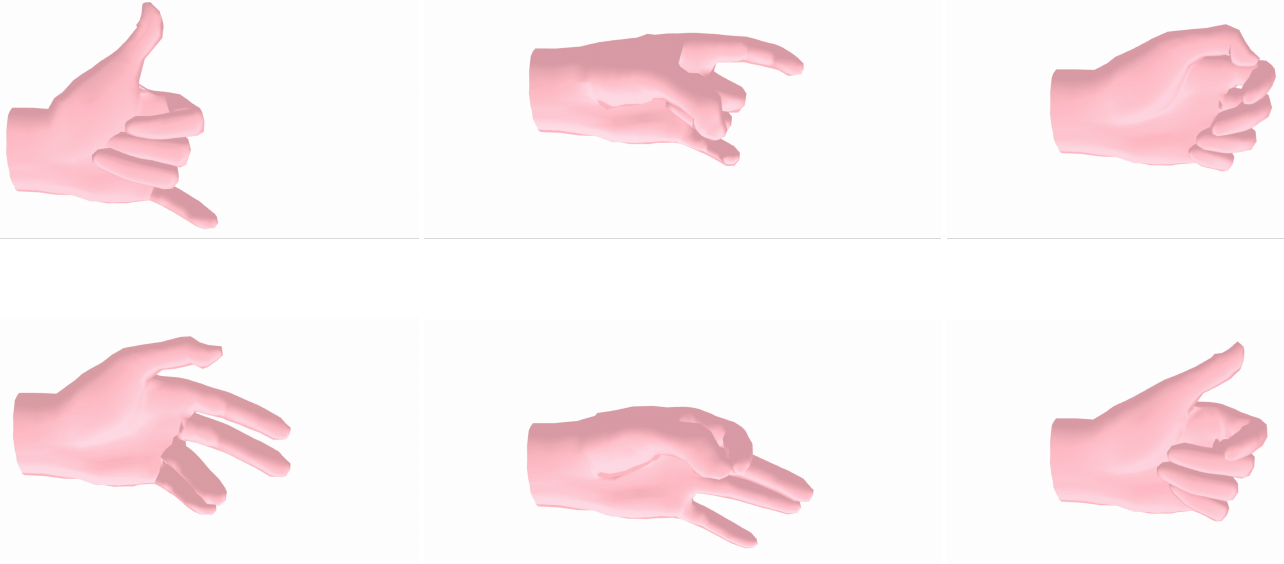}
    \caption{Example visualizations of gestures used in gesture classification tasks.}
    \label{fig:vis_gesture}
\end{figure}
\section{Implementation Details}

We use 2\,s windows sampled at 2\,kHz for both pose and EMG. EMG is instance-normalized, band-pass filtered (2–250\,Hz), and notch-filtered at 60\,Hz. Following \cite{salter2024emg2poselargediversebenchmark}, we apply channel-rotation augmentation to EMG. 
Our MAE is an encoder–decoder Transformer model with 4 encoder layers and 2 decoder layers, and the embedding dimension is $d{=}256$. We optimize with AdamW~\cite{loshchilov2019decoupledweightdecayregularization} (lr $1\mathrm{e}{-4}$, weight decay $1\mathrm{e}{-5}$) and cosine annealing with warm restarts~\cite{loshchilov2017sgdrstochasticgradientdescent}. The masking ratio is 50\%. Token length is $S{=}200$ for pose and $S{=}50$ for EMG, producing non-overlapping tokens along time. Batch size is 256, and each MAE is trained for 100 epochs. Our supervised baseline PoseT shares the same transformer architecture with MAE model, the only difference is that the outputs are pose regression predictions. The training objective is also the same as MAE, a mean squared error loss.

For CPEP, we attach a 1-layer projection head (hidden size 256) to the EMG encoder and train the EMG encoder plus projection head while keeping the pose encoder frozen. The contrastive temperature $\tau$ is learnable (initialized to 0.02). Batch size is 256 and training runs for 100 epochs. All output embeddings are $\ell_2$-normalized. All model trainings are conducted on $4\times$ NVIDIA V100 GPUs; end-to-end training of each model takes approximately 4.5 hours. 
In zero-shot retrieval, we precompute pose embeddings for the entire corpus and, for each EMG query, retrieve the top-$k$ neighbors by cosine similarity with $k{=}10$; the predicted label is the majority vote of the retrieved labels.

\end{document}